\pdfoutput=1

\documentclass[11pt]{article}

\usepackage[preprint]{acl}

\usepackage{times}
\usepackage{latexsym}
\usepackage{amsmath}
\usepackage{amsfonts}  
\usepackage{adjustbox}
\usepackage{nicefrac}  
\usepackage[T1]{fontenc}

\usepackage[utf8]{inputenc}
\usepackage{algorithm}
\usepackage{algorithmic}

\usepackage{microtype}

\usepackage{inconsolata}

\usepackage{graphicx}
\usepackage{booktabs}

\title{CPTQuant - A Novel Mixed Precision Post-Training Quantization Techniques for Large Language Models}

\author{Amitash Nanda \\
  UC San Diego, ECE \\
  La Jolla, CA, USA \\
  \texttt{ananda@ucsd.edu} \\\And
  Sree Bhargavi Balija \\
  UC San Diego, ECE \\
  La Jolla, CA, USA \\
  \texttt{sbalija@ucsd.edu} \\\And
  Debashis Sahoo \\
  UC San Diego, CSE \\
  La Jolla, CA, USA \\
  \texttt{dsahoo@ucsd.edu} \\}

\begin{document}
\maketitle
\begin{abstract}

Large language models have transformed the comprehension and generation of natural language tasks, but they come with substantial memory and computational requirements. Quantization techniques have emerged as a promising avenue for addressing these challenges while preserving accuracy and making energy efficient. We propose CPTQuant, a comprehensive strategy that introduces correlation-based (CMPQ), pruning-based (PMPQ), and Taylor decomposition-based (TDMPQ) mixed precision techniques. CMPQ adapts the precision level based on canonical correlation analysis of different layers. PMPQ optimizes precision layer-wise based on their sensitivity to sparsity. TDMPQ modifies precision using Taylor decomposition to assess each layer’s sensitivity to input perturbation. These strategies allocate higher precision to more sensitive layers while diminishing precision to robust layers. CPTQuant assesses the performance across BERT, OPT-125M, OPT-350M, OPT-1.3B, and OPT-2.7B. We demonstrate up to 4x compression and a 2x-fold increase in efficiency with minimal accuracy drop compared to Hugging Face FP16. PMPQ stands out for achieving a considerably higher model compression. Sensitivity analyses across various LLMs show that the initial and final 30\% of layers exhibit higher sensitivities than the remaining layers. PMPQ demonstrates an 11\% higher compression ratio than other methods for classification tasks, while TDMPQ achieves a 30\% greater compression ratio for language modeling tasks. 

\end{abstract}

\section{Introduction}

\begin{figure}[h]
  \includegraphics[width=\columnwidth]{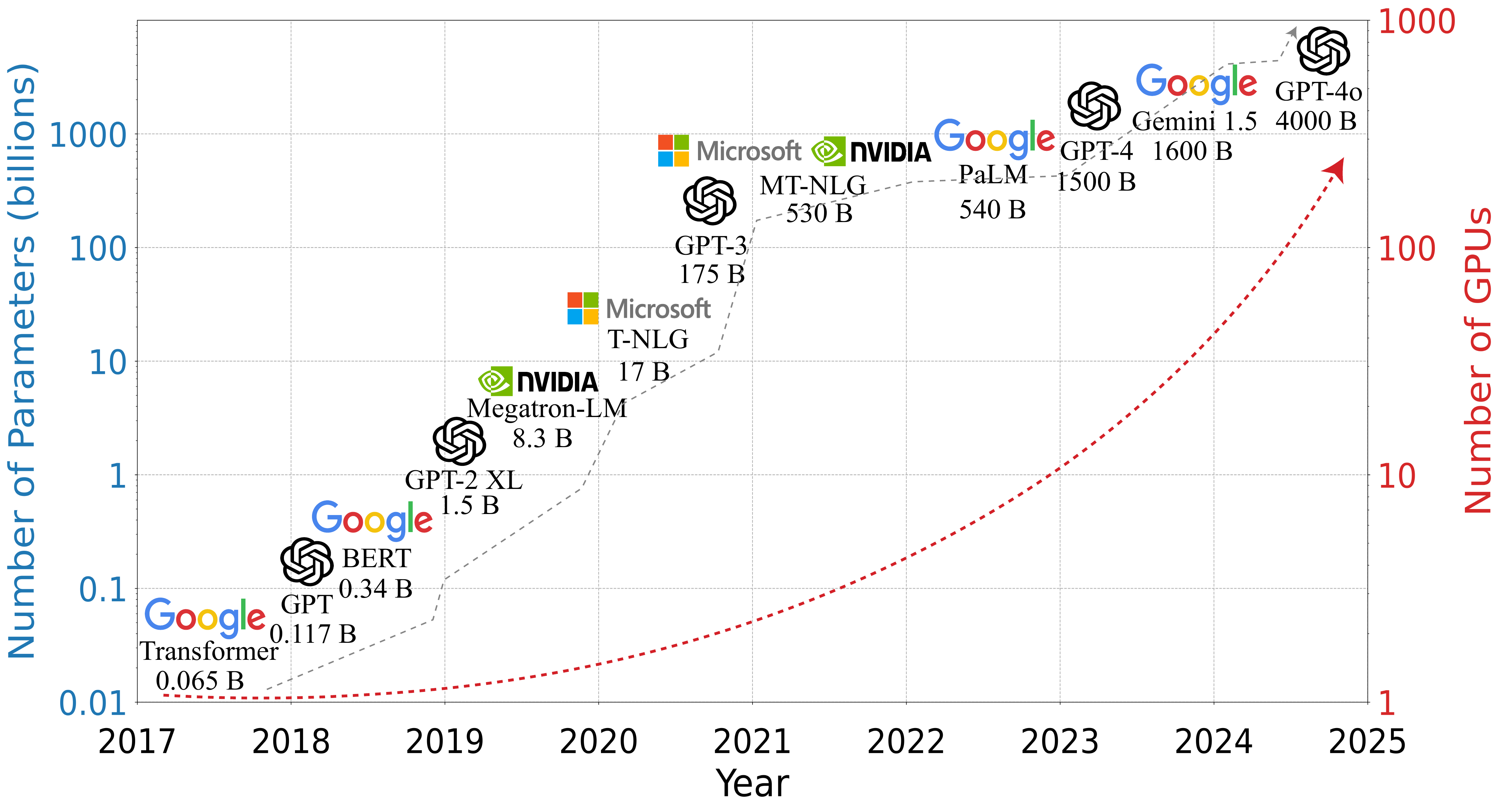}
  \caption{Visualization of Comparision of LLMs: Parameters and GPU requirement increases by 10x.}
  \label{fig:1}
\end{figure}

Large Language Models (LLMs) like GPT, Gemini, Llama, etc., \citep{brown2020language,team2023gemini,touvron2023llama,zhang2022opt} have demonstrated ground-breaking advancement in a variety of applications \citep{wu2023autogen,stiennon2020learning,chen2023llm,balija2024building} in understanding and modeling natural language tasks. However, achieving such exemplary performances involves training trillions of parameters, leading to larger model sizes but higher model quality \citep{hoffmann2022training,kaplan2020scaling} as shown in Figure~\ref{fig:1}. For example, the GPT- 4 model \citep{achiam2023gpt} contains approximately 1 trillion parameters, consuming at least 2TB of memory to store and run in FP16 with 25x80 GB A100 GPUs for inference. The extensive size illustrates the model's complexity and the necessary computational resources. Fine-tuning LLMs for downstream tasks \citep{wei2021finetuned} adapts a pre-trained model to perform specialized tasks using additional training. By leveraging the knowledge acquired in pre-training, the fine-tuning step enables models to achieve high performance on various applications. However, fine-tuning a large-scale language model with billions or even trillions of parameters \citep{fedus2022switch} is computationally intensive. Therefore, several parameters and memory-efficient fine-tuning strategies have been introduced \citep{houlsby2019parameter,kim2024memory} for less memory storage and task-specific parameter updates during deployment. Methods like LoRA reduce memory usage during fine-tuning; for example, GPT-4 still requires 350 GB of storage for parameters in FP16 after fine-tuning. Despite the remarkable efficacy of LLMs, the financial and energy demands of the same pose significant challenges while scaling or deploying. Therefore, a considerable focus has been on compressing weights and activation for LLMs using techniques like pruning and quantization \citep{frantar2023sparsegpt,santacroce2023matters,ma2023llm,lin2023awq,frantar2022gptq,kim2023squeezellm}.

 So, quantization has emerged as a favorable method for reducing memory size, preserving accuracy, and making the model energy efficient. Moreover, the process involves storing the model parameters at a lower precision than the 32-bit or 16-bit used for training purposes. One of the effective solutions is post-training quantization (PTQ); this method significantly reduces training prerequisites and simultaneously lowers the weights to lower precisions INT8 or INT4. Post-training quantization reduces the model size and speeds up the inference time, making it feasible to deploy in resource-constrained environments. Unfortunately, post-training quantization below 8-bit often leads to substantial accuracy loss, and in some instances, even higher numerical precision may be necessary. This paper aims to overcome this limitation by effectively utilizing all the information encoded in the pre-trained model and calibration set.

 To tackle the aforenoted challenges, we strive to develop an optimal quantization strategy for contemporary hardware, which typically supports 16, 8, and 4-bit data types with per-channel quantization of weights. Our approach involves a three-stage pipeline that employs techniques on a small calibration set to calculate the sensitivities of different layers. This is followed by integer programming to optimize the bit-width allocation across different layers, thereby reducing overall accuracy loss. Our method adapts mixed-precision and is less susceptible to overfitting than existing approaches, achieving top-notch results for 8-bit quantization on OPT- 1.3B and BERT-base models trained on the IMDB and WikiText datasets, respectively \citep{maas-EtAl:2011:ACL-HLT2011,merity2016pointer}. This paper presents several innovations in mixed-precision post-training quantization, including developing novel algorithms for dynamic precision allocation based on layer sensitivity analysis and integrating Taylor decomposition techniques for enhanced accuracy after quantization. These advancements not only reduce computational overhead but also maintain or even improve the accuracy of the models when deployed in resource-constrained environments. CPTQuant makes sure to serve large language models like Opt-1.3B and Opt-2.7B using only half the GPUs compared to FP16. Our package makes large language models (LLMs) more accessible by offering a comprehensive solution that reduces operational costs. We anticipate that CPTQuant will stimulate further research in this area and can be a step toward making these models available to a broader audience. Our contributions are (i) CPTQuant, an innovative framework for mixed precision post-quantization training that utilizes non-uniform quantization. (ii) Initially, we determine the sensitivities of the model's various layers using our method and assign precision levels based on each layer's sensitivity. (iii) We assess the framework by measuring the accuracy drop after quantization. (iv) Through comprehensive experiments on different LLMs, we demonstrate that our method sets a new benchmark for post-training mixed precision quantization performance.

\section{Related Works}
There have been many approaches in post-training quantization in the literature, but the effectiveness of PTQ has been underscored in many studies \citep{yao2022zeroquant,frantar2022gptq,dettmers2023case}. Moreover, the study of post-training mixed precision quantization of Large language models still needs to be explored. Consequently, developing an effective, hardware-compatible, and ideally training-free mixed precision quantization approach for LLMs that addresses all compute-intensive operations must still be solved. In the literature, there has been significant effort in quantization during training \cite{courbariaux2015binaryconnect,han2015deep,zhou2017incremental,lin2023awq}. These methods provide strategies to speed up inference through quantization and compensate for model degradation. One of the research \citep{leviathan2023fast} increases the inference time for transformers and involves an approach to handle queries with varied latency constraints effectively. Moreover, it involves a unique acceleration technique called speculative decoding for faster inference.

Post-training quantization is a more straightforward technique applied after the model is fully trained, making it easier and faster to deploy. However, in such scenarios, if quantization is not strategically implemented, it can lead to significant accuracy degradation \citep{frantar2022optq,krishnamoorthi2018quantizing,jacob2018quantization}. In the GPTQ study \citep{frantar2022gptq}, the quantization is applied exclusively to model weights, ignoring the activations and leveraging the inference speedups. Recent methodologies in the literature aim to balance model performance with computational efficiency. For instance, Zeroquant implements a per-token quantization \citep{yao2022zeroquant}. This method, designed specifically for LLMS, requires specialized CUDA kernels and has primarily been tested on models with up to fewer parameters. Despite these efforts, maintaining performance comparable to larger models remains challenging. In another approach, Gpt3.int8() \citep{dettmers2022gpt3} combines INT8 and FP16 to address activation outliers. Though this method controls data range, it can introduce latency overheads and possibly making less efficient than using FP16 alone. To address activation outliers, the outlier suppression technique \citep{wei2022outlier} uses non-scaling LayerNorm and token-wise clipping. These methods are effective for smaller models such as BERT \citep{devlin2018bert} and BART \citep{lewis2019bart} but struggle to maintain accuracy in larger LLM configurations.

Researchers have begun exploring cost-effective techniques for larger LLM models to facilitate efficient inference. SmoothQuant \citep{xiao2023smoothquant} enables 8-bit quantization for both weights and activations and significantly reduces memory usage and computational demands. The activation-aware weight quantization (AWQ) \citep{lin2023awq} method selectively protects salient weights based on activation observation. Half precision (FP16) optimizes the performance of neural networks by using 16-bit floating point precision, significantly reducing memory usage and speeding up computation compared to full precision (FP32). Additionally, LUT-GEMM \citep{park2022lut} introduces efficient GPU kernels tailored for specific binary-coding-based quantization. Though several post-training quantization schemes are available in the literature, mixed-precision post-training quantization methodologies are relatively rare. Our proposed approach utilizes mixed-precision post-training quantization and demonstrates more sophisticated and precise strategies to quantize large-language models. Specifically, CPTQuant achieves more than double the compression compared to previous techniques while maintaining a similar level of accuracy.

\section{Method}

\subsection{Problem Setup}

Consider a trained network $M$ with $L$ layers and trained weights $W_L$. To represent the weights in a designated integer format using $b$ bits (e.g., \texttt{int8} or \texttt{float16}), we use a quantization operator $Q$. This operator transforms the range $[\min\{W_l\}; \max\{W_l\}]$ to the quantized interval $[-2^{b-1}; 2^{b-1} - 1]$ on the integer scale $\mathbb{Z}$. The quantization involves applying a scaling factor $\text{scale} (s)$ and rounding off the scaled tensor. Let $S_L$ be the sensitivities obtained from the CPTQuant package. The $L$ layers of the network are categorized into three distinct groups, $L1$, $L2$, and $L3$, based on their respective magnitudes. Layers with the highest sensitivities are allocated 16-bit precision, those with moderate sensitivities receive 8-bit precision, and those with the lowest are assigned 4-bit precision.

\subsubsection{Quantization}

The quantization function is defined as follows:
\begin{equation}
Q(x) = \Bigg\lfloor \frac{x - \min(x)}{\text{scale}} \Bigg\rfloor + q_{\text{min}}
\end{equation}
where $x$ is the weight matrix to be quantized, $\text{scale} = \frac{\max(x) - \min(x)}{q_{\text{max}} - q_{\text{min}}}$, $q_{\text{min}}$ and $q_{\text{max}}$ are the minimum and maximum quantization levels, $\lfloor \cdot \rfloor$ represents rounding to the nearest integer. \( M_{\text{O}} \) represents the total original memory. \( M_{\text{Q}} \) represents the total quantized memory. Final reduction percentage (FPR) and compression ratio (CR) is defined as follows:
\begin{equation}
\text{FPR} = 100 \times \left( 1 - \frac{M_{\text{O}}}{M_{\text{Q}}} \right)
\end{equation}
\begin{equation}
\text{CR} = \frac{M_{\text{Q}}}{M_{\text{O}}}
\end{equation}

\subsubsection{Objective}

$Q(w)$ represents the quantization function applied to the weights $w$. $L(w, D)$ is the loss function of the model, where $D$ is the dataset. $R(w, Q(w))$ is a regularization term that measures the quantization effect, the norm of the difference between original and quantized weights. $\lambda$ is a regularization parameter that controls the trade-off between the loss minimization and the quantization effect. The optimization problem is formulated using $\arg\min$ as follows:

\begin{equation}
\hat{w} = \arg\min_w \left(A + \lambda B\right)
\end{equation}
\begin{equation}
A = L(Q(w), D)\quad,\quad B = R(w, Q(w))
\end{equation}

This formulation balances loss function minimization while maintaining perplexity and promotes significant quantization of the weights with a greater compression ratio.

\subsection{Correlation-based mixed precision quantization (CMPQ)}

\begin{figure}[h]
  \includegraphics[width=\columnwidth]{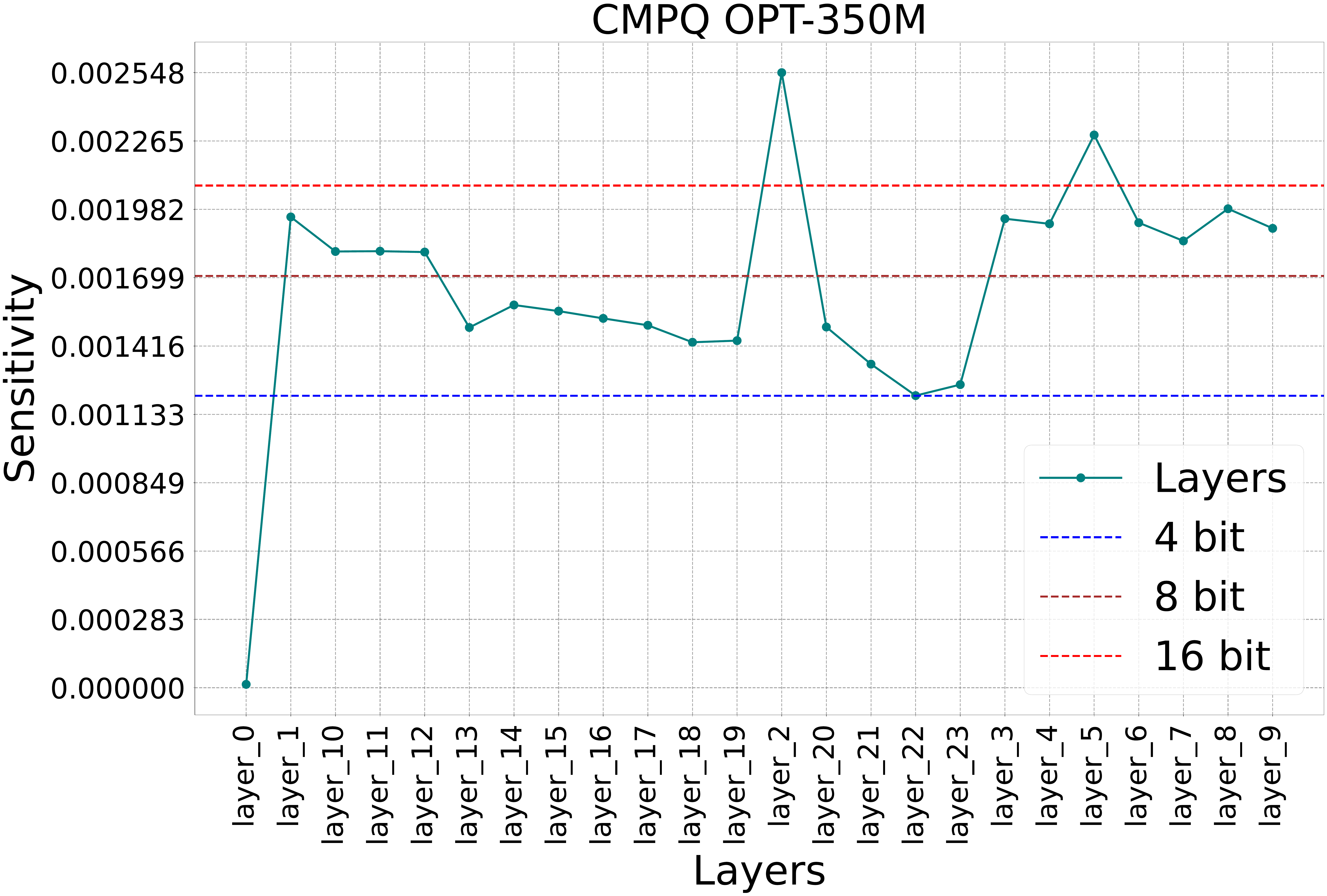}
  \caption{Layerwise sensitivities distribution using the CMPQ method.}
  \label{fig:2}
\end{figure}

Correlation-Based Mixed Precision Quantization (CMPQ) is our first innovative approach to optimizing large language models. This technique uses canonical correlation analysis (CCA) to assess the sensitivity of each layer in a model by examining the correlation between different layers. By measuring how changes in one layer affect other layers, CMPQ can determine which layers are most sensitive to alterations and, consequently, require higher numerical precision during quantization. As explained in Algorithm~\ref{algo:1}, CMPQ first tokenizes and passes data through an LLM to extract outputs from each layer. These outputs are then analyzed using CCA to establish a correlation profile for each layer relative to others. Layers with lower correlations are highly sensitive and are assigned higher precision (16-bit) to preserve their computational integrity and minimize information loss after quantization. Conversely, layers with higher correlations are less sensitive and quantized to lower precisions (8-bit or 4-bit) without significant loss of functionality. Leveraging K-means clustering as shown in Figure~\ref{fig:2}, we categorize the sensitivity of different LLM layers into three distinct groups and assign appropriate precision levels accordingly. A detailed explanation of CCA is shown in Appendix~\ref{sec:appendix}.

\begin{algorithm}
\caption{CMPQ Algorithm}
\label{algo:1}
\begin{algorithmic}[1] 
\STATE Load model, tokenizer, dataset $\rightarrow$ Define quantized model, $Cr$, Accuracy Drop.
\FOR{each layer $i$ in number of layers}
    \STATE Sensitivity using CCA $\rightarrow$ Calculate mean sensitivity, output.
\ENDFOR
\FOR{each layer $i$}
    \STATE Precision Sensitivities $\rightarrow$ Quantized weights.
\ENDFOR
\STATE Evaluate model accuracy pre and post-quantization.
\end{algorithmic}
\end{algorithm}

\subsection{Pruning-based mixed precision quantization (PMPQ)}
\begin{figure}[]
  \includegraphics[width=\columnwidth]{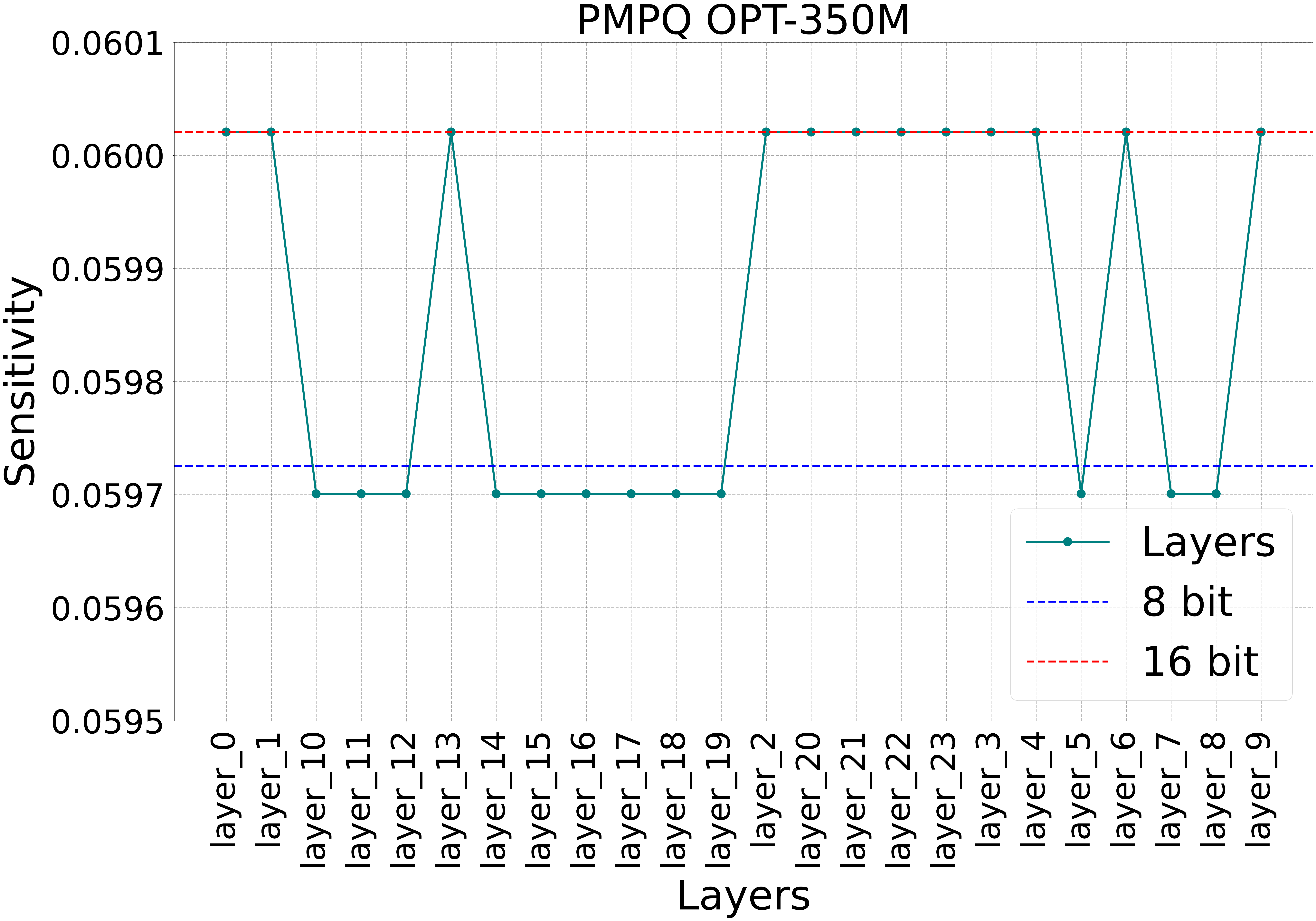}
  \caption{Layerwise sensitivities distribution using the PMPQ method.}
  \label{fig:3}
\end{figure}

Pruning-Based Mixed Precision Quantization (PMPQ) is our second innovative approach to optimize the efficiency and performance of large language models by intelligently varying the precision of quantization across different layers based on their sensitivity to sparsity. As explained in Algorithm~\ref{algo:2}, this method begins with evaluating a baseline model's accuracy on a specific task, such as a language modeling task, using a comprehensive dataset like WikiText for benchmarks. Subsequently, the model undergoes a systematic alteration where each encoder layer of an OPT model is pruned independently to a predetermined sparsity level to assess its impact on the model's accuracy. By leveraging the insights gained from sensitivity analysis as shown in Figure~\ref{fig:3}, PMPQ aims to achieve an optimal balance between model size, speed, and accuracy. The final model is then rigorously evaluated to confirm that the performance metrics, such as classification accuracy and language modeling perplexity, meet the desired standards. This method provides a path toward more scalable and efficient AI systems, particularly in environments where computational resources are at a premium. Among these three methods, PMPQ has demonstrated outstanding performance by compressing the model 4X while only experiencing a minimal accuracy drop of 0.3. PMPQ would be an excellent method to integrate with NVIDIA TensorRT-LLM for categorization tasks.

Applying sparsity in neural networks involves generating a mask based on the weight magnitudes relative to a predefined threshold, where \( w_i \) are the layer weights.

\begin{algorithm}[t]
\caption{PMPQ Algorithm}
\label{algo:2}
\begin{algorithmic}[1] 
\STATE Load model, dataset.
\STATE Initialize data loader and device $\rightarrow$ Evaluate base accuracy.
\FOR{each sparsity level $s$}
    \FOR{each layer $l$ in OPT model}
    \STATE  Clone model $\rightarrow$ Apply PMPQ to layer $l$ with sparsity $s$.
    \STATE Evaluate model accuracy.
    \ENDFOR
    \STATE Compute sensitivity $\rightarrow$ Base accuracy - Current accuracy
    \STATE Output layer $l$ sensitivity.
\ENDFOR
\end{algorithmic}
\end{algorithm}
The mask and threshold is determined by:
\begin{equation}
\text{mask}_i = 
\begin{cases} 
1 & \text{if } |w_i| > \text{threshold} \\
0 & \text{otherwise}
\end{cases}
\end{equation}
\begin{equation}
\text{threshold} = \text{quantile}(|w|, \text{sparsity level})
\end{equation}
Here, \( w \) is the flattened weight tensor of a layer, and the sparsity level is the quantile used to compute the threshold. The accuracy of a model is calculated as the average of correctly predicted labels over all batches:
\begin{equation}
\text{Accuracy} = \frac{1}{N} \sum_{i=1}^N (\hat{y}_i == y_i)
\end{equation}
where \( N \) is the total number of batches, \( \hat{y}_i \) are the predicted labels, and \( y_i \) are the true labels. The comparison results in a boolean value that's averaged over all batches.

\subsection{Taylor Decomposition-based Mixed Precision Quantization (TDMPQ)}
\begin{figure}[h]
  \includegraphics[width=\columnwidth]{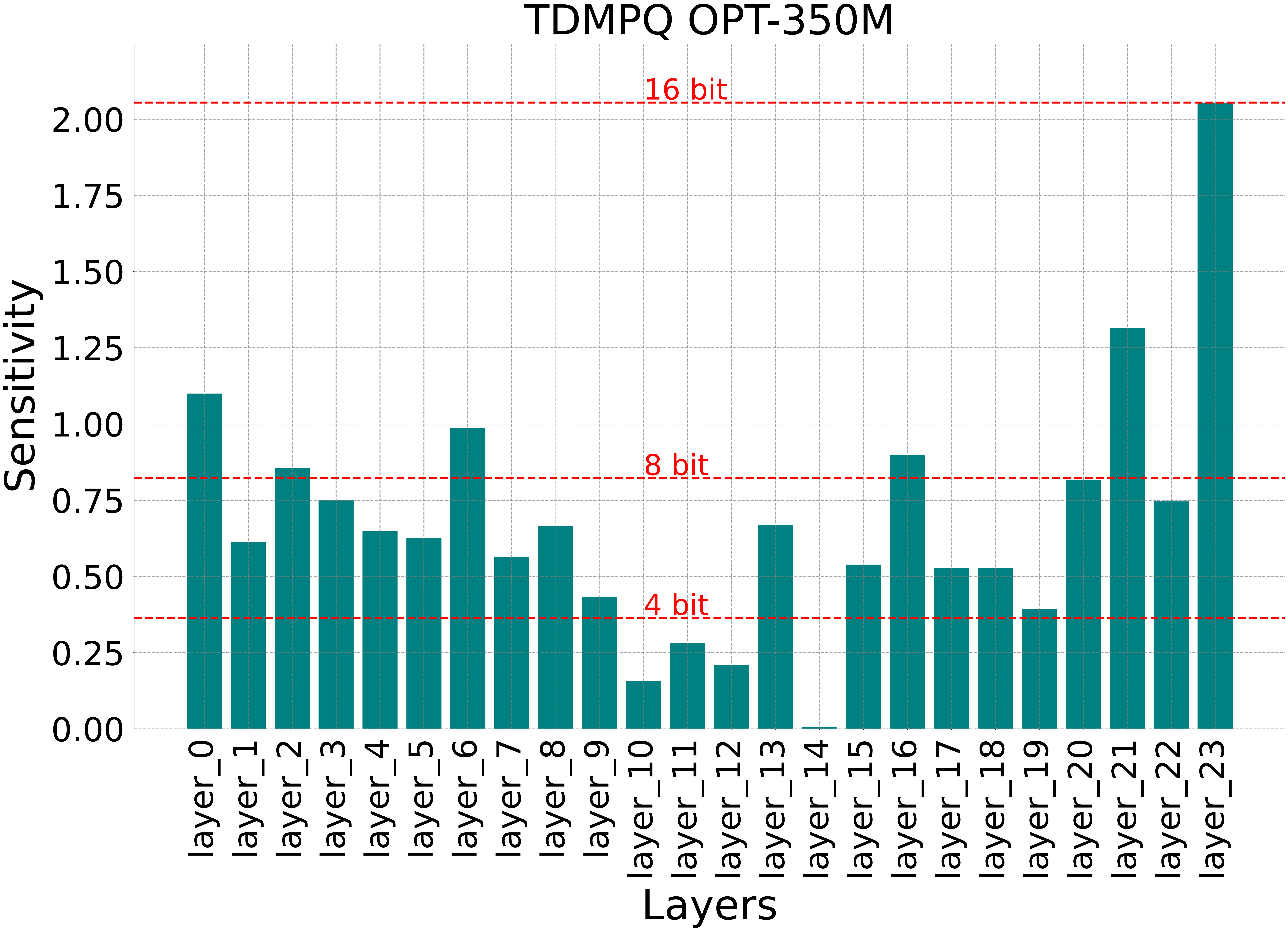}
  \caption{Layerwise Sensitivities Distribution using the TDMPQ Method.}
  \label{fig:4}
\end{figure}

Taylor Decomposition-based Mixed Precision Quantization (TDMPQ) is our third innovative approach that enhances the computational efficiency and performance of large language models like OPT (Open Pre-trained Transformers) through selective precision quantization as explained in Algorithm~\ref{algo:3}. This method leverages Taylor's decomposition to assess the sensitivity of each layer within the model to small perturbations in its inputs, which serves as a basis for applying mixed precision quantization strategies effectively. The primary focus is on calculating the first-order derivatives of the output concerning the inputs. By measuring how the output of each layer responds to these perturbations, we determine the sensitivity of that layer to changes in its inputs. Layers that exhibit higher sensitivity are considered crucial for maintaining the model's performance and are thus assigned higher quantization precision (e.g., 16-bit). Conversely, as shown in Figure~\ref{fig:4}, layers with lower sensitivity, demonstrating robustness to input variations, are quantized at lower precision levels (e.g., 4-bit or 8-bit), reducing the computational resources required without significantly impacting the overall accuracy. Perturbation is applied to the weights as follows:
\begin{equation}
W'_{\text{param}} = W_{\text{param}} + \epsilon
\end{equation}
where \(W'_{\text{param}}\) is the perturbed weight, \(W_{\text{param}}\) is the original weight of the first parameter of the layer, and \(\epsilon\) is the perturbation vector sampled from a normal distribution with the same dimensions as \(W_{\text{param}}\). After perturbation, the total variation (TV) in loss is calculated as:
\begin{equation}
\text{TV} = \sum_{\text{batch} \in \text{Dataloader}} \! L(\text{model}(X_{\text{batch}}))
\end{equation}
where \(L\) represents the loss function, and \(X_{\text{batch}}\) denotes the input batch.

The sensitivity of a layer is computed using the total variation:
\begin{equation}
S_l = \frac{\text{Total Variation}}{N}
\end{equation}
where \(N\) is the total number of samples in the dataset.
After the sensitivity analysis, the original weights are restored to prevent compound modifications across multiple layers:
\begin{equation}
W_{\text{param}} \leftarrow W_{\text{original}}
\end{equation}

\begin{algorithm}
\caption{TDMPQ Algorithm}
\label{algo:3}
\begin{algorithmic}[1] 
\STATE Load model, dataset $\rightarrow$ Initialize data loader on device.
\FOR{each layer $i$ in model}
    \STATE Store original state $\rightarrow$ Perturb first parameter.
    \STATE Compute loss variation across batches $\rightarrow$ Restore original layer state.
\ENDFOR
\STATE Calculate and output normalized sensitivity for each layer.
\end{algorithmic}
\end{algorithm}

\section{Experiments Details}
\subsection{Datasets}

We evaluated our model using two large-scale datasets, WikiText \citep{merity2016pointer} and Imdb \citep{maas-EtAl:2011:ACL-HLT2011}. WikiText is a language modeling dataset with over 100 million tokens extracted from the set of verified goods and featured articles on Wikipedia. IMDB is a binary classification dataset consisting of sentiment data for movie reviews.

\subsection{Baselines and Evaluation Metrics}
We compare our method with the previous state-of-the-art methods on WikiText and IMDb. To evaluate the performance of each method (PMPQ, CMPQ, TDMPQ), we use the three standard metrics: Compression ratio (Cr), Accuracy drop (Ad), and Perplexity Drop (Pd). A higher compression ratio with a lesser accuracy drop indicates better performance.

\begin{figure}[t]
  \includegraphics[width=\columnwidth]{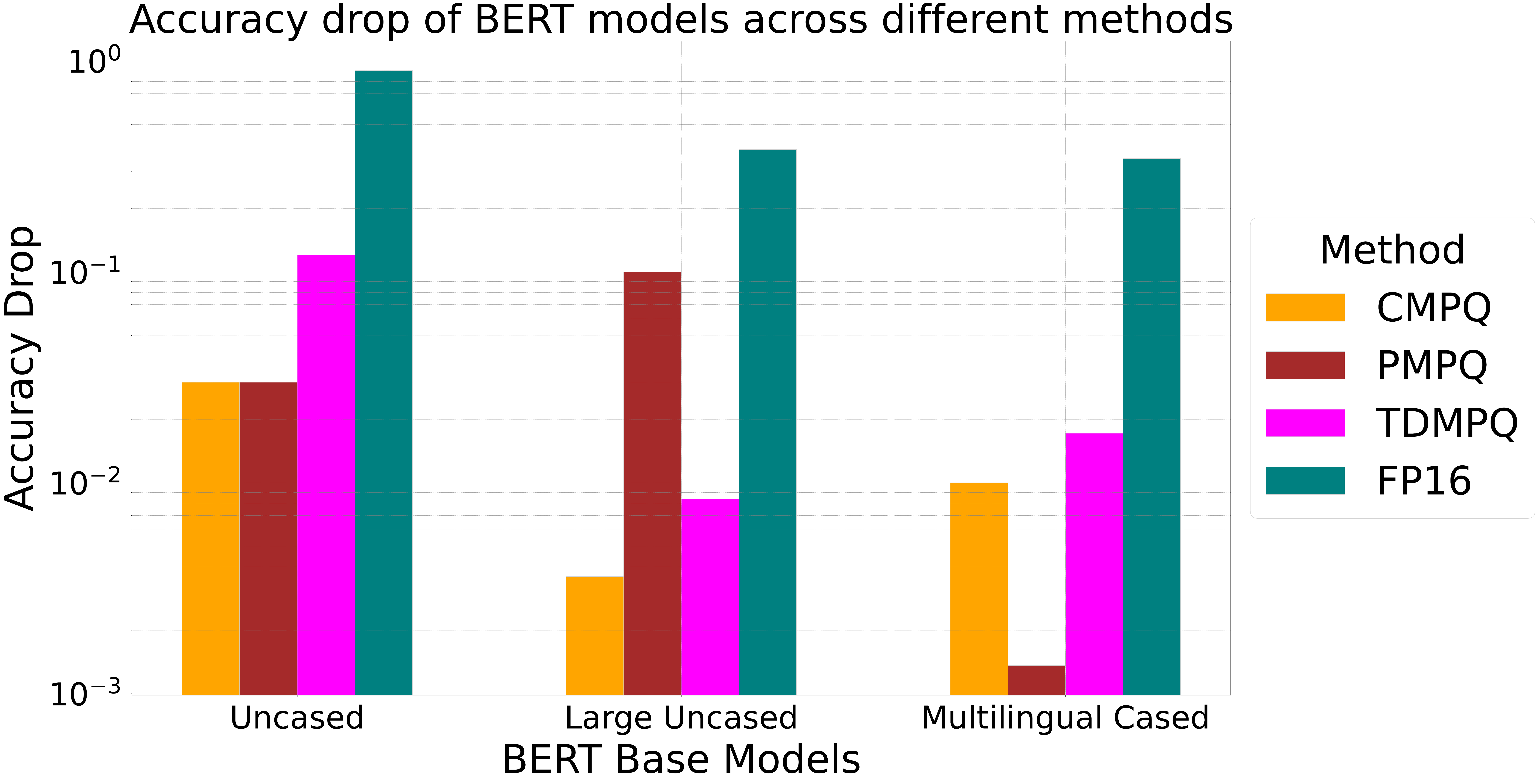}
  \caption{Comparision of accuracy drop of different types of BERT models using CMPQ, PMPQ, TDMPQ with FP16.}
  \label{fig:5}
\end{figure}

\begin{figure}[h]
  \includegraphics[width=\columnwidth]{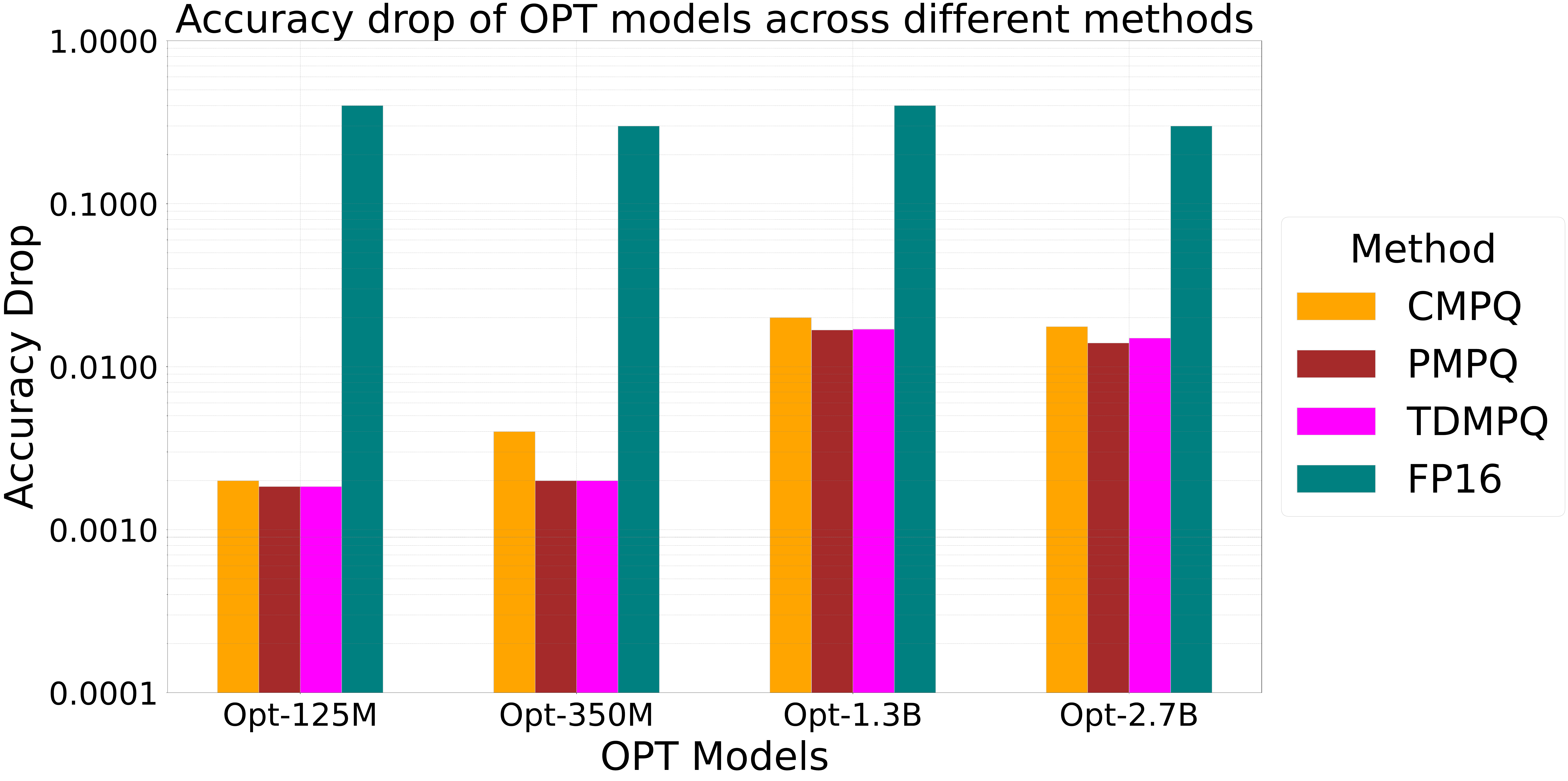}
  \caption{Comparision of accuracy drop of different types of OPT models using CMPQ, PMPQ, TDMPQ with FP16.}
  \label{fig:6}
\end{figure}

\begin{figure*}[t]
  \includegraphics[width=0.48\linewidth]{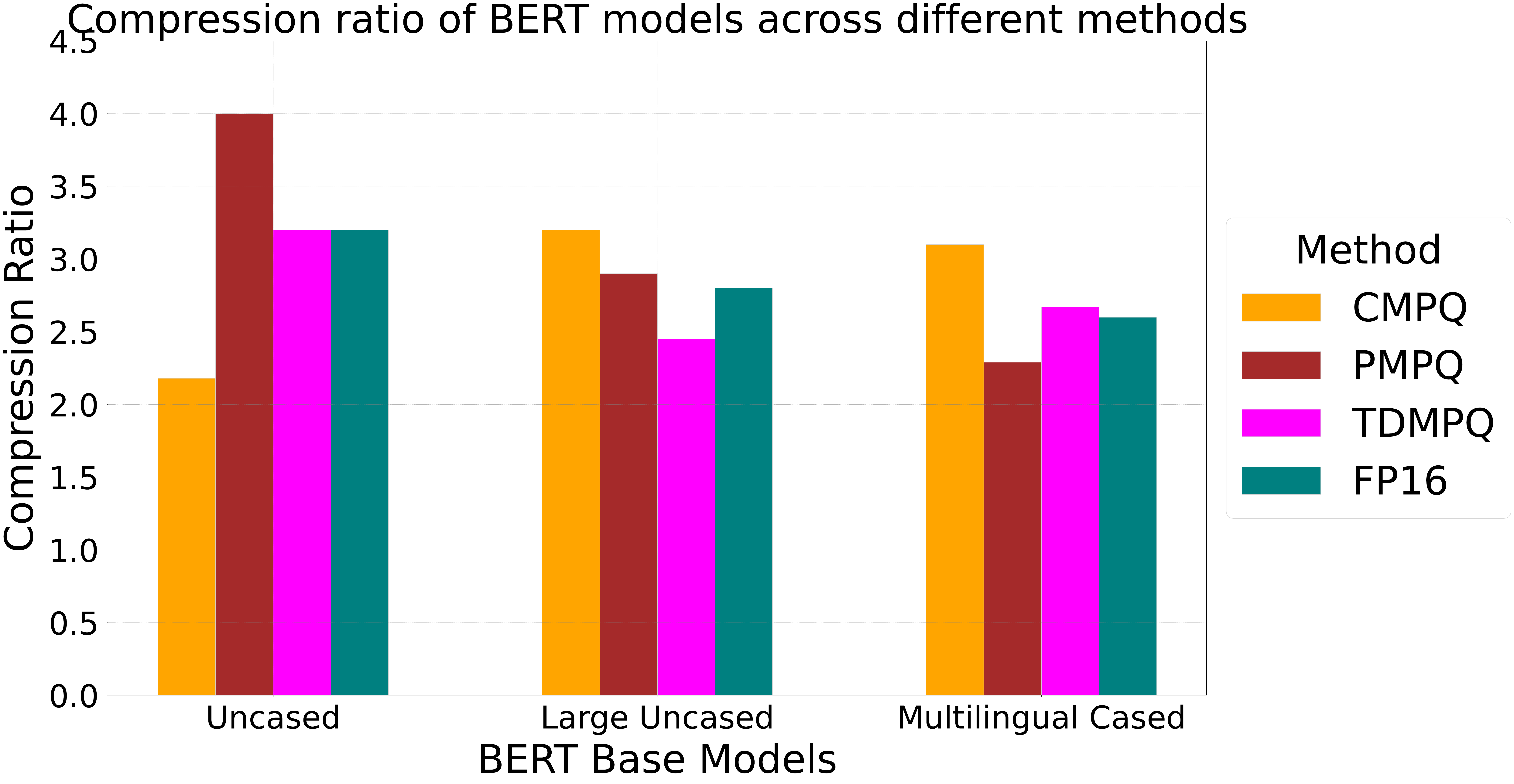} \hfill
  \includegraphics[width=0.48\linewidth]{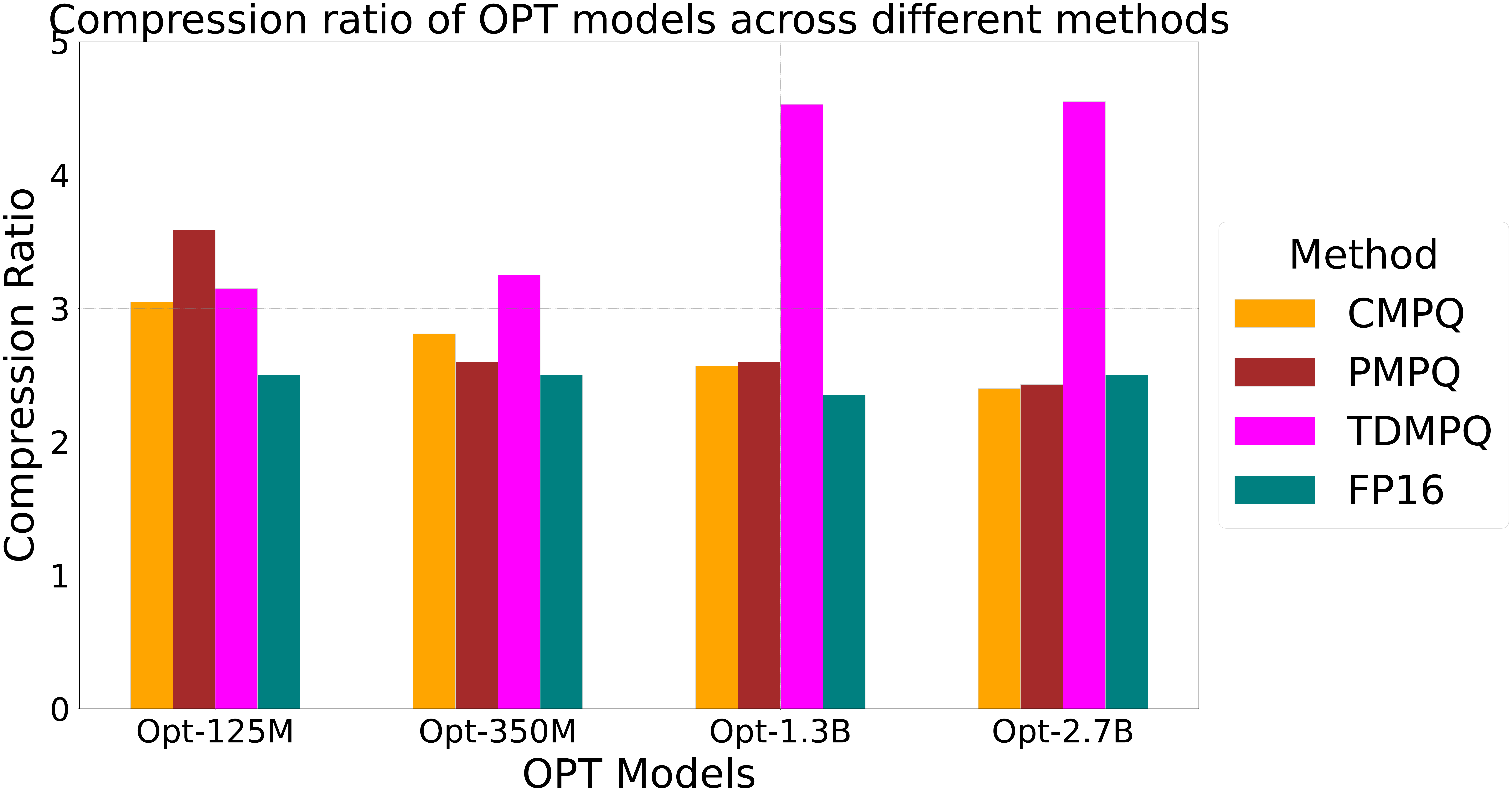}
  \caption {Comparision of the compression ratio of different types of BERT and OPT models using CMPQ, PMPQ, TDMPQ with FP16.}
  \label{fig:7}
\end{figure*}

\begin{table*}[h]
  \centering
  \begin{tabular}{lccc}
    \hline
     \textbf{Model} & \textbf{OPT 125M} & \textbf{OPT 350M} & \textbf{OPT 1.3B}\\ 
     \hline
    First 30\% Layers  & 3.573     & 4.108    & 7.681          \\
    Mid 30\% Layers    & 3.183     & 3.451    & 5.724       \\
   Remaining Layers   & NaN     & 3.662    & 3.662         \\
    \hline
  \end{tabular}
  \caption{Average Standard Deviation from Mean Sensitivity across different OPT Model sizes (125M, 350M, 1.3B, 2.7B), segmented by first 30\%, middle 30\%, and remaining layers.}
   \label{tab:1}

\end{table*}

\begin{figure}[h]
  \includegraphics[width=\columnwidth]{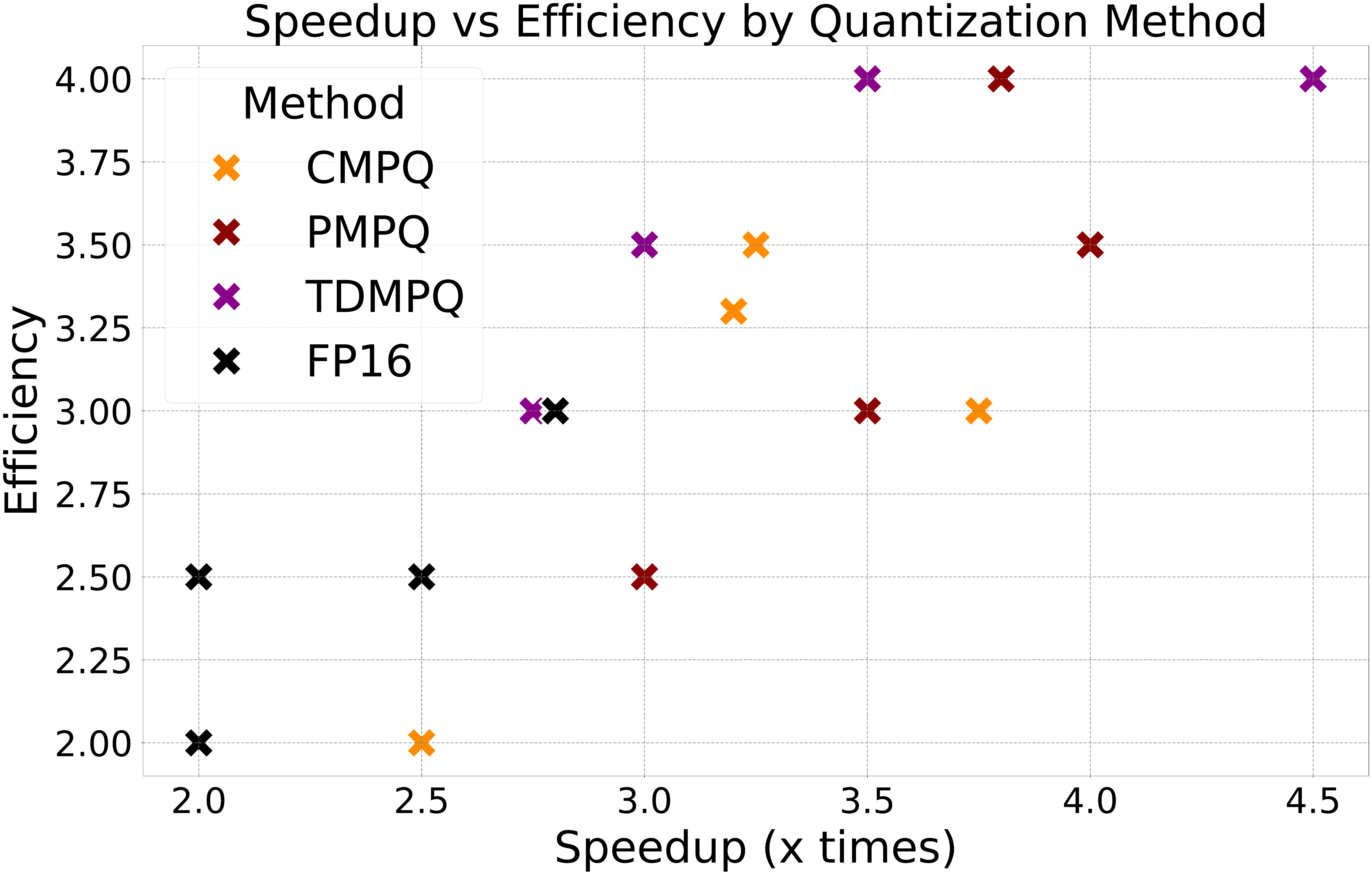}
  \caption{Comparision of speed and efficiency of CMPQ, PMPQ, TDMPQ with FP16.}
  \label{fig:8}
\end{figure}

\subsection{Experimental Setup and Results}
Our experiments used Amazon SageMaker, leveraging instances optimized explicitly for machine learning tasks. To execute the OPT-1.3B and OPT-2.7B models, we utilized the g4dn.12xlarge instance, which provided the necessary computational power and memory to train and test our models efficiently. Amazon SageMaker enabled scalable deployment and facilitated the management of computational resources, ensuring consistent performance throughout our experiments. A detailed explanation of the hardware used and results is shown in Appendix~\ref{sec:appendix1}.

\subsection{Superior Performance of our Quantization Methods Over FP16}
The methods in CPTQuant consistently show lower accuracy drops compared to the FP16 method across several BERT and OPT models. This indicates CPTQuant's higher effectiveness in maintaining the model's performance post-quantization. This is crucial for applications where preserving the model's accuracy is vital, such as tasks requiring high reliability and precision. In models like OPT-1.3B, CMPQ exhibits an accuracy drop of just 0.02 compared to FP16's more significant drop of 0.4, demonstrating CMPQ's superior ability to maintain model precision under quantization as shown in Figure~\ref{fig:5} and Figure~\ref{fig:6}. Table \ref{tab:1} shows different OPT models with average standard deviation from mean sensitivity segmented by first 30\%, middle 30\%, and last remaining layers.


\subsection{Increased Compression Ratios}
Figure~\ref{fig:7} results show that this method maintains better accuracy and provides higher compression ratios than FP16. This suggests that these methods are more efficient in reducing model size without compromising much on performance. Higher compression ratios are beneficial for deploying models on devices with limited storage and processing capabilities, such as mobile devices and embedded systems. TDMPQ stands out by achieving a compression ratio of 4.53 in the Opt-1.3B model on the WikiText dataset, which is significantly higher than FP16's ratio of 2.35, underscoring TDMPQ's efficiency in data reduction while preserving essential model characteristics.

\subsection{Model-Specific Quantization Suitability}
Figure~\ref{fig:8} and other results indicate that the effectiveness of a quantization method can vary significantly between different models. For example, some strategies that work well with OPT-350M might perform less effectively with OPT-2.7B. This highlights the importance of selecting a quantization method tailored to each model's specific characteristics and requirements, ensuring optimal performance and efficiency. Despite the high compression ratios, PMPQ in the OPT-2.7B model keeps the perplexity drop to a minimal five on the WikiText dataset, far better than the ten observed with FP16, indicating a solid balance between compression and performance retention. The detailed comparison in Table~\ref{tab:2} of all the model performances with our three strategies and the FP16 benchmarked model with IMDB and WikiText data summarises the efficiency of CPTQuant. 

\begin{table*}[t!]
\centering

\begin{tabular}{|c|c|c|c|c|c|c|}
\hline
\textbf{Model} & \textbf{Method} & \multicolumn{2}{c|}{\textbf{IMDB}} & \multicolumn{2}{c|}{\textbf{WikiText}} \\
\cline{3-6}
 &  & \textbf{Accuracy Drop} & \textbf{Cr} & \textbf{Perplexity Drop} & \textbf{Cr} \\
\hline
BERT base model  & CMPQ & 0.03 & 2.18x & 5 & 3.019x \\
 & PMPQ & 0.03 & 4x & 4 & 3.21x \\
 & TDMPQ & 0.12 & 3.2x & 8 & 3.644x \\
 & FP16 & 0.9 & 3.2x & 12 & 2x \\
\hline
BERT large model  & CMPQ & 0.0036 & 3.2x & 2 & 3.055x \\
 & PMPQ & 0.1 & 2.9x & 7 & 3.45x \\
 & TDMPQ & 0.0084 & 2.45x & 6 & 3.7x \\
 & FP16 & 0.38 & 2x & 12 & 2x \\
\hline
BERT multilingual base model & CMPQ & 0.01 & 3.1x & 10 & 3.33x \\
 & PMPQ & 0.00136 & 2.29x & 5 & 2.17x \\
 & TDMPQ & 0.0172 & 2.67x & 7 & 3.85x \\
 & FP16 & 0.345 & 2x & 12 & 2x \\
\hline
OPT-125M & CMPQ & 0.002 & 3.05x & 6 & 2.91x \\
 & PMPQ & 0.00184 & 3.59x & 6 & 3.89x \\
 & TDMPQ & 0.00184 & 3.15x & 3 & 2.86x \\
 & FP16 & 0.4 & 2.5x & 12 & 2x \\
\hline
OPT–350M & CMPQ & 0.004 & 2.81 & 7 & 4.33x \\
 & PMPQ & 0.002 & 2.60x & 6 & 3.85x \\
 & TDMPQ & 0.002 & 3.25x & 8 & 3.14x \\
 & FP16 & 0.3 & 2.5x & 10 & 2x \\
\hline
OPT-1.3B & CMPQ & 0.02 & 2.57x & 7 & 4.33x \\
 & PMPQ & 0.01681 & 2.60x & 8 & 3.85x \\
 & TDMPQ & 0.017 & 4.53x & 9 & 3.14x \\
 & FP16 & 0.4 & 2.35x & 12 & 2x \\
\hline
OPT-2.7B & CMPQ & 0.0176 & 2.4x & 6 & 4.25x \\
 & PMPQ & 0.014 & 2.43x & 5 & 3.88x \\
 & TDMPQ & 0.015 & 4.55x & 4 & 3.34x \\
 & FP16 & 0.3 & 2.5x & 10 & 2x \\
\hline
\end{tabular}
\caption{Comparison of model performance across CMPQ, PMPQ, TDMPQ, FP16 using IMDB and WikiText dataset using accuracy drop, compression ratio, and perplexity drop.}
\label{tab:2}
\end{table*}

\section{Conclusion}
 In this paper, we propose CPTQuant, a package of three novel mixed precision quantization techniques that surpass the constraints of existing approaches by diminishing the complexity of implementation while enhancing the model's compressibility with minimal reduction in perplexity. We demonstrate that CPTQuant outperforms existing state-of-the-art post-training quantization methods in accuracy and computational efficiency. The PMPQ method achieves an 11\%  higher compression ratio than other methods in grouping tasks, whereas TDMPQ attains a 30\% more excellent compression ratio in language modeling tasks. Additionally, we provide CMPQ, PMPQ, and TDMPQ for convolution and transformer versions, respectively, to demonstrate the scheme's satisfactory architecture generality. The larger model (OPT-1.3B) consistently shows higher standard deviations from the mean sensitivity than the smaller models (OPT-125M and OPT-350M) across all segments. This suggests that larger models may have layers with more varied sensitivities, and this is due to more complex or diverse representations learned by larger models or potentially more specialized layers that react differently depending on the specific function they serve in the model.
From the analysis, we consider prioritizing CMPQ and PMPQ for broader use across various NLP models. Considering their generally lower error rates and competitive performance metrics, further optimizations might be necessary for TDMPQ, particularly in handling complex models like Llama-7B and OPT-2.7B.

\section*{Acknowledgments}
We thank all the reviewers and mentors who provided valuable insights into our work. We also sincerely thank Bilge Acun (Meta) for giving feedback on our methods and their scope for varied LLM applications. We thank Dr. Song Han for the helpful discussions at ASPLOS. We are grateful to Dr. Debashis Sahoo for constructive feedback on an early draft of this paper.

\section*{Limitations}
Our experiments were limited to publicly available datasets. Testing our current methods on large-scale language modeling datasets will provide valuable insights. Due to computational challenges, we couldn't test our strategies on large-scale LLM models like Llama 2 7B, 13B, and 70B. In our future work, we plan to extend this work to large vision models like VILA-2.7B and language models like Llama-3 and Gemini 1.5 and further aim to implement targeted fine-tuning stages post-quantization. This will enable the model to adjust effectively to the modified head configurations by employing strategies such as differential learning rates on underperforming data segments. Then, the model can better adapt to these changes. These fine-tuning enhancements are designed to mitigate any potential accuracy declines resulting from the quantization of the heads, thereby enhancing the model's overall performance. 

\section*{Ethical Impact}
We have used publicly available datasets to assess the performance of each strategy proposed in this research across different open-source pre-trained LLM models. Our research benchmarked various parameter sizes of the LLM model (from small to large) with Hugging Face FP16. Through this comprehensive study, we could generalize our strategies and compare accuracy drop and compression ratio. CPTQuant addresses the environmental impact of large language models involving compute-intensive tasks. The proposed methodologies will help make LLMs energy efficient while preserving accuracy and making such large models to deploy efficiently to resource-constrained environments.


\appendix
\section*{Appendix}
\section{Methods}
\label{sec:appendix}
\subsection{Canonical Correlation Analysis (CCA)}
Canonical Correlation Analysis (CCA) solves a specific optimization problem to identify linear combinations of features from different layers outputs that are maximally correlated. The correlation coefficient obtained through this method is crucial for understanding the sensitivity or dependency of one layer's outputs on another. This insight is particularly valuable for exploring the internal dynamics of neural networks, offering a deeper look at how different layers interact and influence each other's behavior.

Find \( \mathbf{w}_X \) and \( \mathbf{w}_Y \) to maximize \( \text{corr}(\mathbf{X} \mathbf{w}_X, \mathbf{Y} \mathbf{w}_Y) \), where:
\begin{itemize}
    \item \(\mathbf{X}\) and \(\mathbf{Y}\) are the feature matrices from two different layers,
    \item \(\mathbf{w}_X\) and \(\mathbf{w}_Y\) are the weight vectors to be found,
    \item \(\text{corr}(\cdot, \cdot)\) denotes the correlation function.
\end{itemize}
Maximize:
\begin{equation}
\mathbf{w}_X^\top \mathbf{C}_{XY} \mathbf{w}_Y
\end{equation}
Subject to:
\begin{equation}
\mathbf{w}_X^\top \mathbf{C}_{XX} \mathbf{w}_X = 1 \quad \text{and} \quad \mathbf{w}_Y^\top \mathbf{C}_{YY} \mathbf{w}_Y = 1
\end{equation}

where:
\begin{itemize}
    \item \(\mathbf{C}_{XY}\) is the covariance matrix between \(\mathbf{X}\) and \(\mathbf{Y}\),
    \item \(\mathbf{C}_{XX}\) and \(\mathbf{C}_{YY}\) are the covariance matrices of \(\mathbf{X}\) and \(\mathbf{Y}\) respectively.

\end{itemize}

\begin{figure}[t]
  \includegraphics[width=\columnwidth]{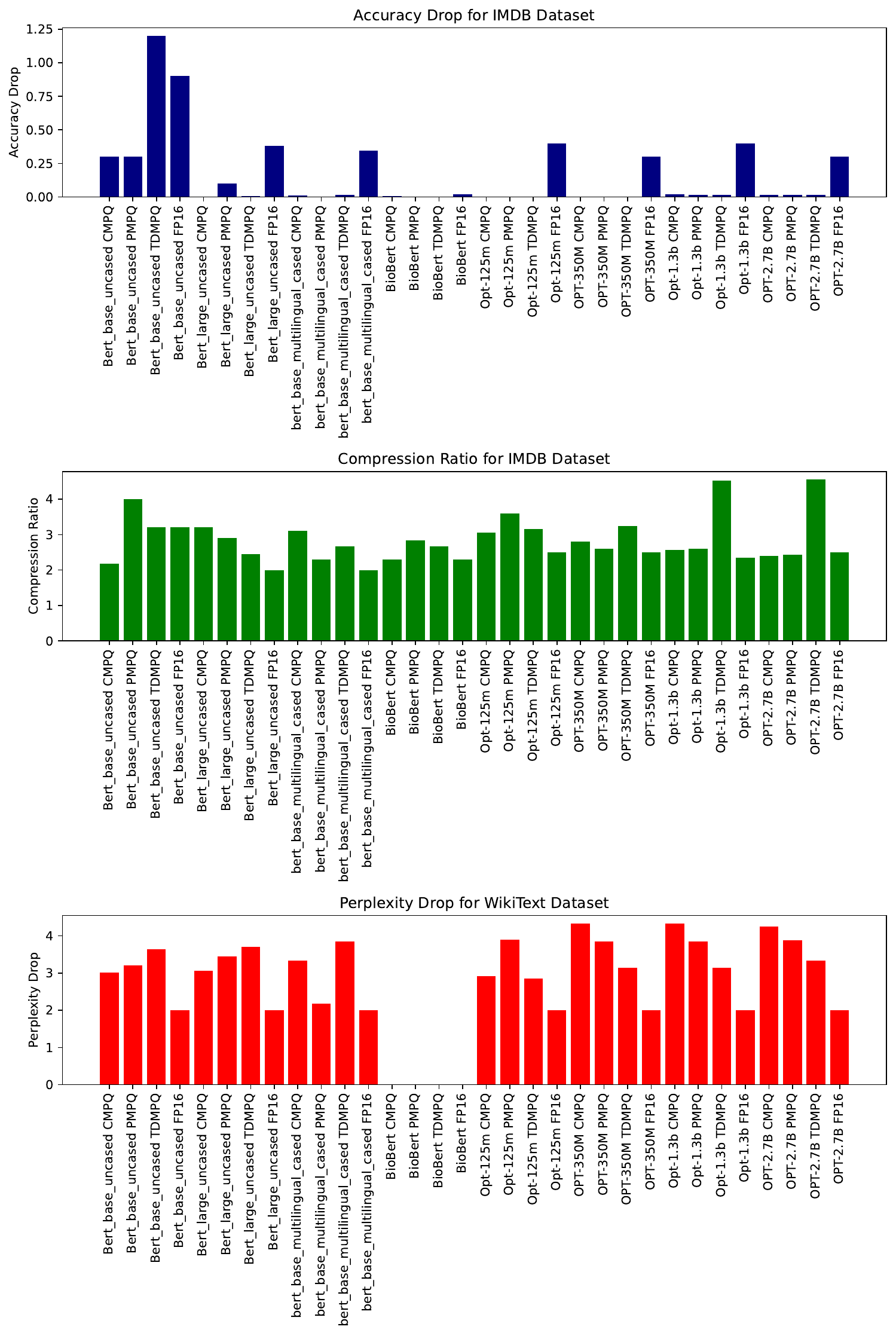}
  \caption{Accuracy Drop, Compression ratio, and Perplexity drop for IMDB and WikiText data across all models.}
  \label{fig:9}
\end{figure}

\section{Experimental Settings and Results}
\label{sec:appendix1}
For models like BERT, we used 4 Nvidia GeForce GTX 1080 graphics cards. We also used the PyTorch accelerator package for parallel processing using 4-GPU while training and inference. For large models like OPT, we used Amazon SageMaker  g4dn.12xlarge instance. It has 48 vCPUs, 192.0 Memory (GiB), Intel Xeon Family, a Clock Speed of 2.5 GHz, 4 GPUs, and 64 GB Video Memory. We spent around 200 USD on AWS usage for our entire research work. Figure~\ref{fig:9} shows the detailed results with different metrics.  

\end{document}